\pdfoutput=1

\documentclass[11pt]{article}

\usepackage{acl}

\usepackage{times}
\usepackage{latexsym}
\usepackage{array}
\usepackage{wrapfig}
\usepackage{color}
\usepackage{algorithmic}
\usepackage[linesnumbered,ruled,vlined]{algorithm2e}
\usepackage[utf8]{inputenc}
\usepackage{multirow}         
\usepackage{xspace}
\usepackage{amssymb} 
\usepackage{colortbl}
\usepackage{booktabs}
\usepackage{amsmath}
\usepackage{amsmath}
\usepackage{booktabs}
\usepackage{diagbox}  
\usepackage{colortbl} 
\usepackage{xcolor}
\usepackage[T1]{fontenc}

\usepackage[utf8]{inputenc}

\usepackage{microtype}

\usepackage{inconsolata}

\usepackage{graphicx}

\title{MuAP: Multi-step Adaptive Prompt Learning for Vision-Language Model with Missing Modality}

\newif\ifaclfinal
\aclfinaltrue

\author{Ruiting Dai,Yuqiao Tan,Lisi Mo,Tao He,Ke Qin, Shuang Liang \\
    University of Electronic Science and Technology of China \\ }

\begin{document}
\maketitle
\begin{abstract}
Recently, prompt learning has garnered considerable attention for its success in various Vision-Language (VL) tasks. However, existing prompt-based models are primarily focused on studying prompt generation and prompt strategies with complete modality settings, which does not accurately reflect real-world scenarios where partial modality information may be missing. In this paper, we present the first comprehensive investigation into prompt learning behavior when modalities are incomplete, revealing the high sensitivity of prompt-based models to missing modalities. To this end, we propose a novel \underline{\textbf{Mu}}lti-step \underline{\textbf{A}}daptive \underline{\textbf{P}}rompt Learning (\textbf{MuAP}) framework, aiming to generate multimodal prompts and perform multi-step prompt tuning, which adaptively learns knowledge by iteratively aligning modalities. Specifically, we generate multimodal prompts for each modality and devise prompt strategies to integrate them into the Transformer model. Subsequently, we sequentially perform prompt tuning from single-stage and alignment-stage, allowing each modality-prompt to be autonomously and adaptively learned, thereby mitigating the imbalance issue caused by only textual prompts that are learnable in previous works. Extensive experiments demonstrate the effectiveness of our MuAP and this model achieves significant improvements compared to the state-of-the-art on all benchmark datasets.
\end{abstract}

\begin{figure}[t]
    \includegraphics[width=1\columnwidth]{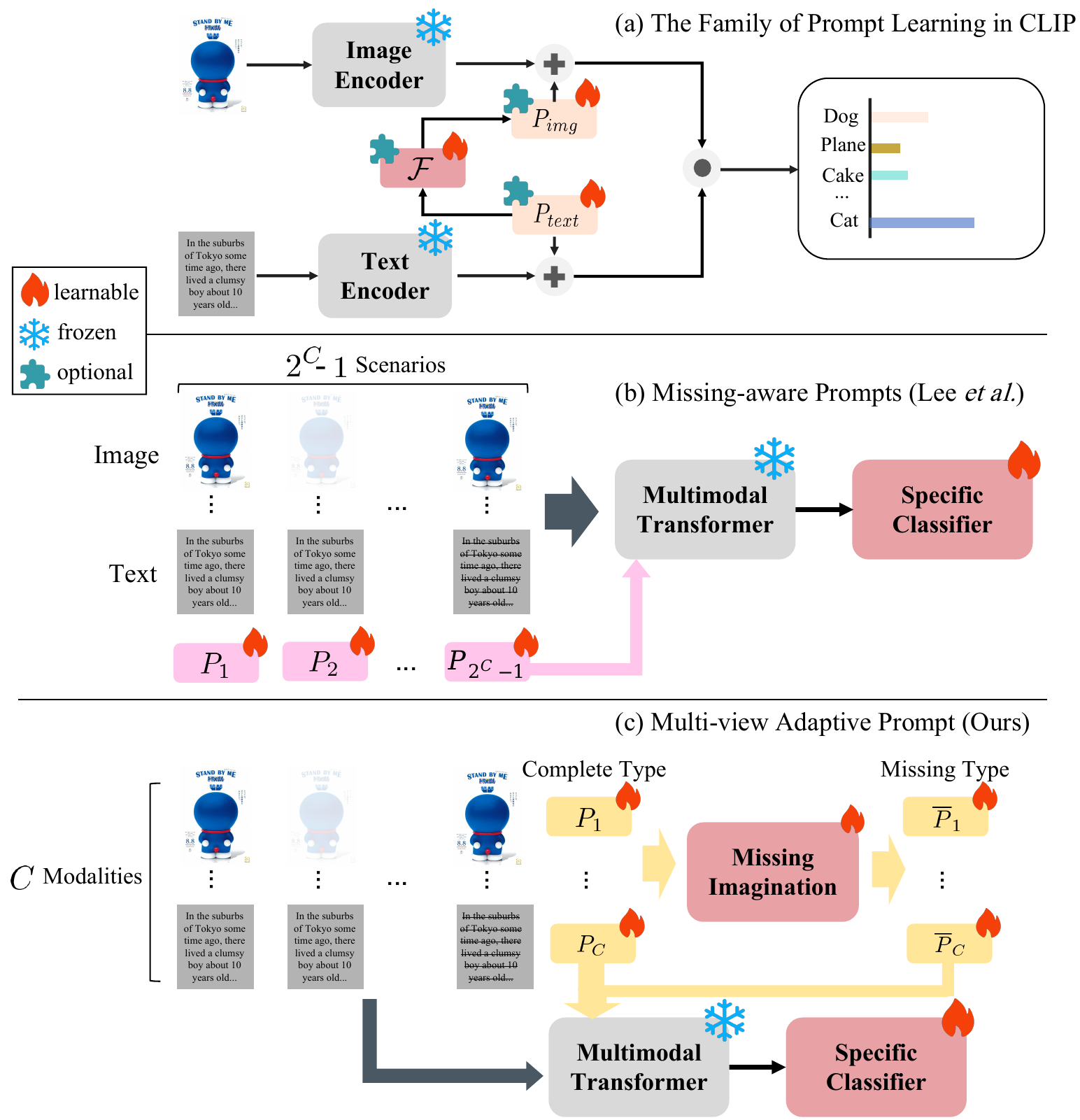}
    \caption{Various architectures in the prompt tuning field. 
    (a) The CLIP-family method \cite{maple} focus on prompt generation with complete modality information. 
    (b) Missing-aware prompts method in MPVR~\cite{map} has $2^C-1$ prompts to represent all missing scenarios, where C is the number of modalities.
    (c) Our method aims to enhance parameter efficiency by utilizing only $C$ prompts and to improve robustness through multi-step prompting tuning in missing scenarios.
    }
    \label{fig:method comparing}
\end{figure}
\section{Introduction}
Vision-Language (VL) pre-training~ \cite{su2019vl,lu2019vilbert,yu2019deep,kim2021vilt} has demonstrated remarkable success in various Vision-Language tasks like image recognition~ \cite{zhang2021tip, liu2019multi}, object detection~ \cite{jin2021cdnet,sun2021deep}, and image segmentation~ \cite{cao2021shapeconv, hu2019acnet} by learning the semantic correlations between different modalities through large-scale image-text training. However, most previous research has assumed that all modalities are accessible during both training and testing phases, a condition that is often challenging to meet in real-world scenarios. This challenge arises from various factors, such as privacy and security concerns leading to the inaccessibility of textual data~ \cite{lian2023gcnet}, or limitations in device observations resulting in missing visual data~ \cite{zeng2022tag,ma2022multimodal}. Hence, the widespread occurrence of missing modalities distinctly hinders the performance of vision-language models.

Recently, as shown in Figure~\ref{fig:method comparing}(a), there has been a notable advancement in the field of visual language  (VL) by adopting prompt learning from Natural Language Processing  (NLP). However, researchers do not consider scenarios where modalities are missing. For instance, CLIP~ \cite{clip} aligns image and language modalities through joint training on large-scale datasets. It leverages handcrafted prompts and a parameterized text encoder to generate precise classification weights, thereby enabling zero-shot learning. Nonetheless, it faces two formidable challenges: the need for expertise and multiple iterations in designing handcrafted prompts, as well as the impracticality of fully fine-tuning the entire model due to its tremendous scale. Consequently, CoOp~ \cite{coop} and CoCoOp~\cite{cocoop} propose automated prompt engineering that converts contextual words in prompts into learnable vectors and achieves substantial improvements by exclusively fine-tuning dense prompts using a small number of labeled images.
Furthermore, MaPLe~ \cite{maple} delves into the limitations of solely using language prompts in previous works and presents multimodal prompt learning, which introduces a coupling function to connect text prompts with image prompts, facilitating mutual gradient propagation between the two modalities for more precise alignment.

Recent research, such as MPVR~ \cite{map}, has proposed using prompt learning for scenarios with missing modalities, aiming to mitigate the performance degradation caused by disparities in modality absence in training or testing data samples. However, designing distinct prompts for each missing modality scenario inevitably leads to an exponential increase in the number of prompts as the number of modalities increases (as shown in Figure~\ref{fig:method comparing}(b), a scenario with $C$ modalities necessitates $2^C-1$ prompts), seriously compromising the scalability of the model. Moreover, unlike the dual-prompt strategy used by MaPLe~ \cite{maple}, MPVR~ \cite{map} adopts a coarse prompt strategy at the input or attention level by directly inserting prompts into multimodal transformers, without distinguishing textual and visual features.

Despite MaPLe 's~ \cite {maple} dual-prompt strategy effectively harnessing the capabilities of both modalities, its coupling mechanism exhibits a propensity for relying predominantly on the textual modality, which may result in unbalanced learning of multimodal information. Furthermore, an excessive degree of coupling has the potential to impede the independent learning capacity of each modality. To address this, in Figure~\ref{fig:method comparing}(c), we propose a novel \underline{\textbf{Mu}}lti-step \underline{\textbf{A}}daptative \underline{\textbf{P}}rompt Learning  (\textbf{MuAP}) framework for multimodal learning in the presence of missing modalities. MuAP introduces a multi-step prompting mechanism that adaptively learns multimodal prompts by iteratively aligning modalities. Specifically, we perform prompt tuning sequentially from two perspectives: single-stage and alignment-stage. This allows each modality prompt to learn autonomously without interference from the other, facilitating an in-depth exploration of each modality in scenarios where certain modalities are missing. Finally, we obtain the downstream classifier results through multimodal prompt learning, where adaptive prompts effectively mitigate imbalanced learning caused by one-way coupling and only textual prompts are learnable in \cite{maple}.

To summarize, this paper makes the following key contributions:

\begin{itemize}
    \item To the best of our knowledge, this paper is the first study to analyze the robustness of prompt learning on  missing modality data. We propose a novel missing-modality in the VL Model model with multi-step adaptive prompt learning, addressing the limitations of previous works and enhancing prompts through autonomous and collaborative learning simultaneously. 

    \item We devise a multi-step tuning strategy that encompasses single-stage and alignment-stage tunings, where we generate visual and language prompts adaptively through multi-step modality alignments for multimodal reasoning. This facilitates comprehensive knowledge learning from both modalities in an unbiased manner.

    \item We conduct extensive experiments and ablation studies on three benchmark datasets. Extensive experiments demonstrate the effectiveness of our MuAP and this model achieves significant improvements compared to the state-of-the-art on all benchmark datasets.
\end{itemize}




\section{Related work}
\subsection{Vision-Language Pre-trained Model}
Recent researches on Vision-Language Pre-training (VLP) aim to learn semantic alignment between different modalities by leveraging large-scale image-text pairs. There are two architectures of the existing VLP methods: single-stream and dual-stream architectures. In single-stream architectures, image and text representations are concatenated at the feature level and serve as input to a single-stream Transformer. For example, VisualBERT \cite{li2019visualbert} concatenates text embedding sequences and image embedding sequences, which are then passed through a Transformer network. Building upon this work. VL-BERT \cite{su2019vl} utilizes OD-based Region Features on the image side and incorporates a Visual Feature Embedding module. Similarly, ImageBERT \cite{qi2020imagebert} follows a single-stream model with OD for image feature extraction while introducing more weakly supervised data to enhance learning performance.
Alternatively, the dual-stream architectures align image-text representations in a high-level semantic space using two separate cross-modal Transformers. For instance, CLIP \cite{clip} and its variants (such as CoOp \cite{coop} and MaPLe \cite{maple}) employ ResNet~\cite{he2016deep} and ViT models as image encoders, while employing Transformers~\cite{vaswani2017attention} as text encoders. Subsequently, they utilize contrastive learning to predict matching scores between each template entity and the current image, with the highest score indicating the image's classification result.

\subsection{Prompt Learning for Vision-Language Tasks}
As the diversity of Vision-Language (VL) tasks poses a challenge for individually fine-tuning large pre-trained models for each task, Prompt Learning emerges as an effective approach to tackle this challenge. It involves freezing the backbone neural network and introducing prompts, which comprise a small number of trainable parameters, to fine-tune the entire model. This allows for the zero-shot or few-shot application of pre-trained models to new VL tasks in a more parameter-efficient manner than training large models from scratch for each task. 
For example, CoOp \cite{coop} incorporates learnable prompts into the language encoder to fine-tune CLIP, while CoCoOp employs conditional prompts to further enhance the model's generalization ability. MaPLe \cite{maple} argues that learning prompts for the text encoder alone in CLIP are insufficient to model the necessary adaptations required for the image encoder. To address this, MaPLe leverages multimodal prompt learning to fully fine-tune the text and image encoder representations, ensuring optimal alignment in downstream tasks. It employs a coupling function to connect the prompts learned in the text and image encoders, with only the text prompts being trainable. 



\section{Method}
In this section, we detail our methodology by presenting a clear problem definition and introducing our proposed MuAP.

\subsection{Problem Definition}
In this work, we study the missing-modality multimodal learning where the presence of missing modalities can occur in both the training and testing phases. For simplicity while retaining generality,  following ~\cite{huang2019image}, we consider a multimodal dataset that contains two modalities: $\mathcal{M}=\{m_t, m_v\}$, where $m_t$ and $m_v$ denote textual, visual modalities respectively. The complete modality data can be represented as $\mathcal{R}^{all}=\{x^{m_t}_i,x^{m_v}_i,y_i\}$, where  $x^{m_t}_i$ and $x^{m_v}_i$ denote the textual and visual features respectively, $y_i$ denotes the corresponding class label. While the missing modality data are $\mathcal{R}^{m_t}=\{x^{m_t}_j, y_j\}$ or $\mathcal{R}^{m_v}=\{x^{m_v}_k,y_k\}$ representing text-only data and image-only data respectively. To keep the format of multimodal inputs, we adopt a straightforward strategy of assigning placeholder inputs, represented as $\overline{x}^{m_t}$ and $\overline{x}^{m_v}$, to the instances with missing modalities. These placeholder inputs are null strings or blank pixels and serve to fill the absence of textual or visual data, respectively.  Consequently, we obtain 
$\overline{\mathcal{R}}^{m_t}=\{x^{m_t}_j,\overline{x}^{m_v}_j,y_j\}$, $\overline{\mathcal{R}}^{m_v}=\{\overline{x}^{m_t}_k,x^{m_v}_k,y_k\}$, and the multimodal data with missing modality can be represented as $\mathcal{R}=\{\mathcal{R}^{all},\overline{\mathcal{R}}^{m_t},\overline{\mathcal{R}}^{m_v}\}$. Our goal is to address classification issues and improve the robustness of VL model with Prompt Learning with missing modalities $\mathcal{R}$.




\begin{figure*}[!t]
    \centering
    \includegraphics[width=0.9\textwidth]{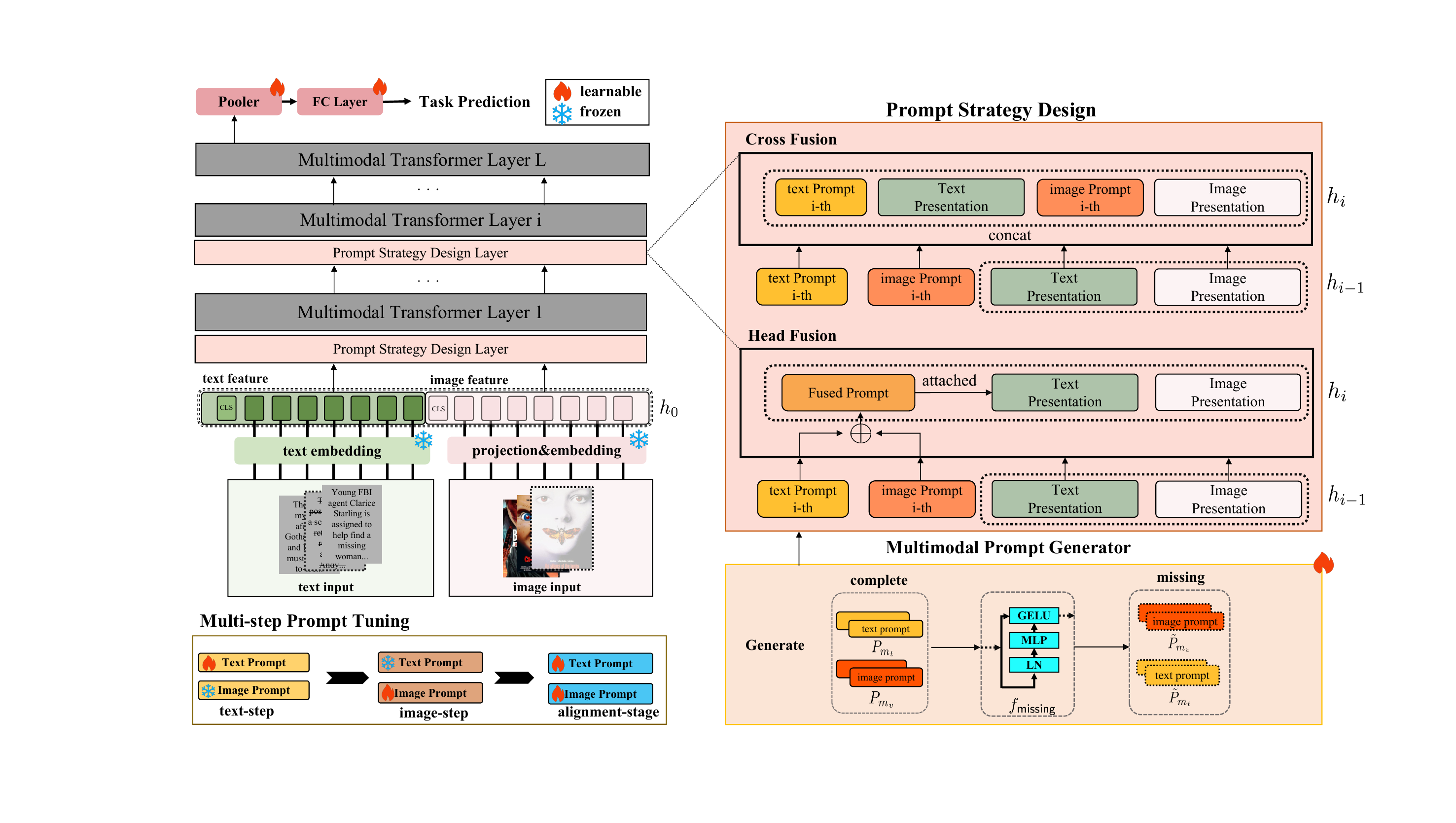}
    \caption{The overview of our MuAP framework. The Multimodal Prompt Generator initially generates complete-type prompts, $P_{m_t}$ and $P_{m_v}$, tailored to the specific modality case (e.g., textual or visual modalities in Vision-Language tasks). Next, it employs $f_\mathsf{missing}$ to create missing-type prompts $\Tilde{P}_{m_t}$ and $\Tilde{P}_{m_v}$ .
    The Prompt Strategy Design module integrates prompts into multiple MSA layers using various strategies (i.e., head fusion or cross fusion). During the training phase, we leverage Multi-step Prompt Tuning to synchronize distinct characteristics of different modality prompts effectively.
   }
\vspace{-2em}
\label{fig:model}
\end{figure*}

\subsection{Overall Framework}
Considering the resource constraints, we focus on the VL model with Prompt Learning and adopt Vision-and-Language Transformer (ViLT)~ \cite{kim2021vilt} as the backbone, which is pre-trained on large-scale VL datasets and remains untrainable in downstream tasks.
To mitigate the significant performance degradation of Prompt Learning models due to missing modality data, we propose a novel \textbf{Mu}lti-step \textbf{A}daptative \textbf{P}rompt Learning (MuAP) model to enhance the model's robustness in various missing scenarios.
As illustrated in Figure \ref{fig:model}, MuAP mainly comprises three modules: Multimodal Prompt Generator, Prompt Strategy Design, and multi-step Prompt Tuning. 
Specifically, we first generate learnable specific prompts for each modality to achieve completeness tuning in prompting, deviating from previous methods~\cite{coop,cocoop} where only textual prompts were learnable. Subsequently, we introduce two prompt fusion strategies: head-fusion and cross-fusion, attaching prompts to blocks of the multimodal transformer. Additionally, we propose a multi-step tuning strategy for dynamic language and vision prompt tuning through modality alignments, allowing MuAP to gain knowledge from both modalities.

\subsection{Revisiting ViLT}
ViLT is a widely used Transformer-based multimodal pretraining model. It partitions images into patches of varying sizes, which are projected and embedded to generate latent representations. This allows the unified processing of images and text with minimal parameters.
Its overall workflow commences by concatenating the text representation (denoted as $t=[t_{cls};t_1;\ldots;t_M]$) with the image patches (denoted as $v=[v_{cls};v_1;\ldots;v_N]$). These concatenated representations are then fed into multiple Transformer layers for processing. Specifically:
\begin{equation}
h^0=[t+t^{modal};v+v^{modal}] \in \mathbb{R^{L_V \times \text{d}}}
\end{equation}
\begin{equation}
\hat{h^i} = \mathsf{MSA}(\mathsf{LN}(h^{i-1})) + h^{i-1},\hspace{0.7em} i=1\ldots L
\end{equation}
\begin{equation}
h^i = \mathsf{MLP}(\mathsf{LN}(\hat{h^{i-1}})) + \hat{h^{i}},\hspace{2em} i=1 \ldots L
\end{equation}
where, $t$ and $v$ represent the embeddings of text and images, respectively. They are combined with their respective modality type embeddings $t^{modal}$ and $v^{modal}$ to form the initial input $h^0$. $L_V$ represents the length of the input sequence, while $d$ denotes the dimension of the hidden states. The context vectors $h$ undergo continuous updates through L layers of Transformer encoders, and the final output context sequence $h^{L}$ is utilized for downstream tasks.

\subsection{Multimodal Prompt Generator}

One main challenge in addressing missing modality learning with prompt learning lies in the design of prompt, and all modality absence situations are exponential. Drawing on the effectiveness of complete prompts in multimodal learning, we generate specific prompts for each modality, with the key distinction being that all the textual and visual prompts are both learnable. Unlike ~ \cite{map}, where missing-aware prompts are generated for each possible situation resulting in an exponential increase as the number of modalities grows, our method adopts a linear growth pattern for prompts that significantly reduces the number of parameters and model complexity. To improve understanding and compensation for missing modalities, we create a simple network to generate specific prompts for each modality, aiding exploration and use of implicit data.

Specifically, when the input comprises $C$ modalities, there exist $C$ complete-type prompts. 
In our VL tasks, given $C=2$ modalities of images and texts, we initialize $P_{m_t}$ and $P_{m_v} \in \mathbb{R}^{L_p \times d}$ as textual and image prompts respectively, representing the complete modality, where $L_p$ is the prompt length.
Subsequently, the initial prompts are fed into a lightweight network $f_{\text{missing}}$, in a crosswise manner. This means that opposing prompts are used to generate prompts (e.g., using a complete-type prompt from the visual modality to generate a missing-type prompt for the textual modality). The goal of this process is to enhance perception and compensate for missing modalities. The formula for the generating process is as follows:
\begin{equation}
    f^i_{\mathsf{missing}} (P^i)= \mathsf{GELU}(\mathbf{W}^i\mathsf{LN}(P^{i})) + P^{i}
\end{equation}
\begin{equation}
    \tilde{P}^i_{m_v} = f^i_{\mathsf{missing}}(P^i_{m_t})
\end{equation}
\begin{equation}
    \tilde{P}^i_{m_t} = f^i_{\mathsf{missing}}(P^i_{m_v})
\end{equation}
where $\mathbf{W}^i$ represents the weight matrix specific to the $i$-th $f_\mathsf{missing}$ module in the $i$-th layer of MSA, $\mathsf{LN}$ refers to the layer normalization operation, $\mathsf{GELU}$ is the activation function, and adding the original prompts $P^i$ represents the residual operation. 
The residual connection is present to retain the opposing modality information while the MLP is utilized to collect additional missing-specific features to provide more valuable supplementary for the missing input and facilitate multimodal fusion. In a more generalized form, let $P_m$ $(m \in \mathcal{M})$ represent the complete-type prompt for modality $m$, and $\Tilde{P}_m$ represent the missing-type prompt for the same modality. When modality $m$ is missing, the missing-type prompt $\Tilde{P}_m$ is utilized in the subsequent module. Otherwise, the complete-type prompt $P_m$ is used.



\subsection{Prompt Strategy Design}
Designing prompt template and strategy is crucial for prompt-based learning. We focus on prompt strategy involving prompt configuration and placement. Two prompt strategies introduced in Figure~\ref{fig:model}: head-fusion prompting and cross-fusion prompting. Consistency in subsequent symbols assumed with complete input data for textual and visual modalities.



\paragraph{\textbf{Head-fusion Prompting.}} One simple way to incorporate prompts is to add them at the start of input sequences for each layer. We use element-wise summation for combining multimodal prompts. $P_{head}$ is expressed as:
\begin{equation}
 P_{head} = P_{m_t} \oplus P_{m_v}, P \in \mathbb{R}^{L_p \times d}
\end{equation}
where $\oplus$ denotes the summation over prompts from each modality. Next, we concatenate $P_{head}$ with the input sequence of texts and images at each layer. Similar to ViLT \cite{kim2021vilt}, the formula can be expressed as follows:
\begin{equation}
 h^{i} = [P_{head}^i;t^i;v^i], \quad i = 0 \cdots N_p
\end{equation}
where $P_{head}^i$ denotes the head-fusion prompt of i-th layer,  $N_p$ represents the number of MSA layers in ViLT. With the concatenating $P_{head}^i$ to the input sequences of the previous layer, the final output length increases to $(N_PL_P+L_V)$ in total. This allows the prompts for the current layer to interact with the prompt tokens inherited from previous layers, enabling the model to learn more effective instructions for prediction.

\paragraph{\textbf{Cross-fusion Prompting.}}
Motivated by ~ \cite{maple}, another prompting approach is to insert modality-specific prompts into their corresponding modality inputs in a single-stream model. By doing this, we facilitate the interaction between modality-specific prompts and features.
The cross-fusion prompting can be formalized as follows:
%
\begin{equation}
    h^i=[P^i_{m_{t}};t^i;P^i_{m_v};v^i], \quad i = 0 \cdots N_p
\end{equation}
where $P^i_{m_{t}}$, $P^i_{m_v}$ represent the modality-specific prompts for the textual and visual modalities, respectively, at the $i$-th layer.
It is noteworthy that, unlike ~ \cite{maple} which only replaces few parameters from the input sequence from each layer, cross-fusion prompt strategy follows head-fusion to attach the prompts at each MSA layer. This results in an expanded final output length of $(2N_PL_P+L_V)$. This improves the model's representation scale and training stability, but it encounters a significant increase in model length when both $N_P$ and $L_P$ are large. It also faces the potential risk of overlooking the information in the original input sequence. We discuss how the prompt length leads to overfitting in Section~\ref{ablation}.

\subsection{Multi-step Prompt Tuning}
In this section, we introduce our proposed multi-step prompt tuning technique designed to adaptively learn multimodal prompts through multi-step sequential modality alignments. Specifically, we employ prompt tuning~\cite{lester2021power} of the pre-trained Transformer encoder to perform efficient parameter learning from multiple stages, including single-stage of each modality and a alignment-stage. This not only facilitates the acquisition of modality-specific information from individual visual and textual modalities but also captures the correlations between different modalities.


\paragraph{\textbf{Single-stage prompt tuning.}} To fully account for the inherent differences between distinct modalities, we sequentially and separately freeze the two modality prompts to explore learnable prompts trained with contrastive learning. As illustrated in Figure~\ref{fig:model}, we iteratively train the learnable prompts in a step-wise manner. Initially, we optimize the textual prompts while keeping the visual prompts frozen, called text-step. Subsequently, we switch to optimizing the visual prompts while fixing the textual prompts, called image-step. This exclusive updating process enables the prompt tuning to capture modality-specific attributes respectively. 

Specifically, in the two steps, we utilize the Kullback-Leibler (KL) divergence as $\mathcal{L}_{kl}$ to measure the distribution difference between text  and visual prompts. Additionally, we incorporate $\mathcal{L}_{cls}$ as a classification loss to facilitate the fusion. 

%
To mitigate overfitting issues caused by prompt engineering, we employ diverse combinations of parameters $\lambda_t$ and $\lambda_v$ in the two steps of prompt updating, which effectively preserves modality-specific information. 
The formulas are as follows:
\begin{equation}
    \label{loss_1}
    \textbf{Text-step}: \mathcal{L}^t_{total} = \mathcal{L}_{cls} + \lambda_t \mathcal{L}_{kl}(P_{m_t},P_{m_v})
\end{equation}
\begin{equation}
    \label{loss_2}
    \textbf{Image-step}: \mathcal{L}^v_{total} = \mathcal{L}_{cls} + \lambda_v\mathcal{L}_{kl}(P_{m_t},P_{m_v})
\end{equation}

During this separate training of modality prompts, the hyper-parameter $\lambda$ is used to combine with the KL loss. Specifically, $\lambda_t$ and $\lambda_v$ are set to 0.4 for the text prompt training step and 0.3 for the image prompt training step, respectively. In the process of single-stage prompt tuning, the two prompts undergo simultaneous updates through several alignment steps, with the experimental setup setting the number of steps to 3.

%

\paragraph{\textbf{Alignment-stage prompt tuning.}} To further adapt multimodal prompts and enhance the generalization capability of downstream tasks, we train the model again from a alignment stage. In this step, the visual and textual prompts are all trainable during the training. The overall training objective solely emphasizes the classification loss $\mathcal{L}_{cls}$, which is formulated as follows:
\begin{equation}
    \textbf{Alignment-stage}: \mathcal{L}_{total} = \mathcal{L}_{cls}
\end{equation}

\section{Experiments}

\subsection{Datasets and Metrics}

\paragraph{\textbf{Datasets}}
We follow the approach outlined in~\cite{map} to evaluate our methods across three multimodal downstream tasks:
\begin{itemize}
    \item \textbf{\textit{MM-IMDb}}~ \cite{mmimdb} focuses on classifying movie genres using both images and text, handling cases where a movie fits into more than one genre.
    \item \textbf{\textit{UPMC Food-101}}~ \cite{food101} is a multimodal classification dataset  and comprises 5\% noisy image-text paired data gathered from Google Image Search. 
    \item \textbf{\textit{Hateful Memes}}~ \cite{hatememes} is a challenging dataset for identifying hate speech in memes through images and text. It has 10k tough samples to challenge unimodal models and favor multimodal models.
\end{itemize}

\paragraph{\textbf{Metrics}} Given the distinct classification tasks addressed by these datasets, we employ appropriate metrics tailored to each dataset. Specifically, for MM-IMDb, we utilize F1-Macro as a measure of multi-label classification performance. For UPMC Food-101, the metric is classification accuracy. For Hateful Memes, we assess performance using the AUROC.

\begin{table*}[!t]
    \centering
    
    \resizebox{\linewidth}{!}{
    \begin{tabular}{c|c|cccc|ccccccc}
    \toprule
    \multirow{2}{*}{Datasets} & 
    \multirow{2}{*}{\shortstack{Missing \\ rate $\epsilon$}} & 
    \multicolumn{2}{c}{Training} &
    \multicolumn{2}{c|}{Testing} & 
    \multirow{2}{*}{ViLT} &
    \multirow{2}{*}{\shortstack{MPVR  \\(Input-level)}} &
    \multirow{2}{*}{\shortstack{MPVR \\ (Attention-level)}} &
    \multirow{2}{*}{\shortstack{Visual \\ BERT~\cite{li2019visualbert}}} &
    \multirow{2}{*}{\shortstack{Ma Model~\cite{ma2022multimodal}}} &
    \multirow{2}{*}{\shortstack{MuAP \\ (Head Fusion)}} &    
    \multirow{2}{*}{\shortstack{MuAP \\ (Cross Fusion)}} \\
    &&Image &Text &Image &Text
    \\ \hline
    \multirow{3}{*}{\shortstack{MM-IMDb  \\ \\ (F1-Macro)}} 
    &\multirow{3}{*}{70\%} 
    &30\% &100\% & 30\% &100\% &37.61 &46.30 &44.74 &38.63&46.63 &\textbf{47.21} & 46.73\\
    &&65\% &65\% & 65\% &65\% &36.30 &42.41 &41.56 &37.23&41.28&42.57 &\textbf{43.92}  \\
    &&100\% &30\% & 100\% &30\% &34.71 &39.19 &38.13&36.41&38.65& \textbf{41.37} & 39.88\\
    
    \hline     

    \multirow{3}{*}{\shortstack{Food101  \\ \\ (Accuracy)}} 
    &\multirow{3}{*}{70\%} 
    &30\% &100\% & 30\% &100\% &76.93 &86.09 &85.89&77.41&86.38&\textbf{86.90} & 86.59 \\
    &&65\% &65\% & 65\% &65\% &69.03 &77.49 &77.55&71.06&78.58&77.87 & \textbf{78.95} \\
    &&100\% &30\% & 100\% &30\% &66.29 &73.85 &72.47&67.78&73.41&\textbf{74.61} & 74.60\\
      \hline     
   
    \multirow{3}{*}{\shortstack{Hateful Memes  \\ \\ (AUROC)}} 
    &\multirow{3}{*}{70\%} 
    &30\% &100\% & 30\% &100\% &61.74 &62.34&63.30&61.98&63.56&65.09 & \textbf{66.83}\\
    &&65\% &65\% & 65\% &65\% &62.83 &63.53 &62.56&63.05&64.41&\textbf{64.76} & 62.68\\
    &&100\% &30\% & 100\% &30\% &60.83 &61.01 &61.77&60.89&60.96&\textbf{62.08} & 61.26\\
    \bottomrule
    \end{tabular}
    }
     \caption{Quantitative results on the MM-IMDB~ \cite{mmimdb}, UPMC Food-101~ \cite{food101}, and Hateful Memes~ \cite{hatememes} with missing rate $\epsilon$\% = $70$\% . The outcomes were analyzed under diverse missing-modality cases, with the best results highlighted in \textbf{bold} for clarity.
    } 
    \label{table:baseline comparison}
    
\end{table*}

\begin{figure*}[!t]
    \centering
    \includegraphics[width=0.9\textwidth]{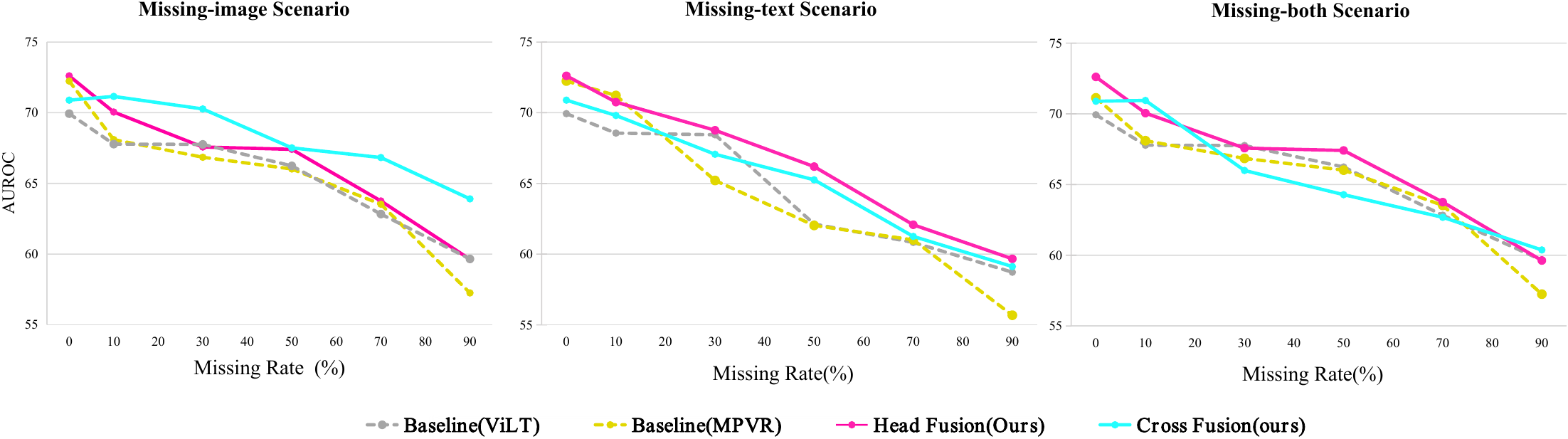}
    \caption{Comparison of baselines on the Hateful Memes dataset with different missing rates across various missing-modality scenarios. Each point in the picture represents training and testing with the same $\epsilon\%$ missing rate.}
    \label{fig:ablation_different_missing_ratio_same_train_test}
\end{figure*}

\subsection{Baselines}

\paragraph{\textbf{Baselines}}  To assess the effectiveness and robustness of our proposed method, we primarily compare it with the state-of-the-art models. These models include
\begin{itemize}
\item \textbf{\textit{Finetuned VILT}}:  the original one without any additional prompt parameters in ViLT (i.e. only training the pooler layer and task-specific classifier).
\item  \textbf{\textit{MPVR~\cite{map}}}: derived from the pre-trained VILT backbone, this model integrates \textbf{missing-aware prompts} into its multimodal transformer design.
\item  \textbf{\textit{Visual BERT~\cite{li2019visualbert}}}: a modified Visual BERT focusing on pooler and classifier training.
\item  \textbf{\textit{Ma Model~\cite{ma2022multimodal}}}: using pre-trained VILT, multi-task optimization, and automated search algorithm to find most efficient fusion technique.
\end{itemize}


\subsection{Main Results}


\paragraph{\textbf{Basic Performance.}}
Table~\ref{table:baseline comparison} shows our new prompt learning method outperforms baselines, demonstrating the effectiveness of our design and training strategy.
The Hateful Memes dataset is tough, making unimodal models struggle, especially with missing modalities. Our head-fusion approach surpasses missing-aware prompts on this dataset, showing a 1.94\% average improvement. This highlights our prompt learning design's proficiency in handling missing data.
Additionally, different fusion strategies lead to distinct modalities integration, with the cross-fusion approach often boosting performance in specific situations, such as when dealing with missing-image cases in the Hateful Memes dataset which surpasses MPVR by about 3.53\%. However, it exhibits greater sensitivity to various missing cases, particularly when text is absent. In scenarios with limited textual data, cross-fusion can inadvertently emphasize the fusion of prompts combined with modality inputs, potentially impacting multimodal representation.

\subsection{Robustness Comparison.}
\paragraph{\textbf{Robustness to Different Missing Rates}}
The performance differences in baseline models vary significantly in robustness to different missing rates. Results for various missing rates on Hateful Memes are displayed in Figure~\ref{fig:ablation_different_missing_ratio_same_train_test}. Assessing robustness involves calculating the average drop rate between successive data points.

MPVR exhibits inferior performance compared to ViLT in certain cases, demonstrating the highest vulnerability with a maximum drop rate of $4.18\%$ in the missing-text scenario and an average drop of $3.53\%$. Our proposed method, compared to head fusion, achieves a significant performance enhancement, with a low drop rate of only $3.05\%$, and average improvements of $9.76\%$ for MPVR and $10.95\%$ for ViLT.
Our cross-fusion strategy demonstrates enhanced performance in most settings of the missing-image scenario, with the lowest drop rate of $2.4\%$. It surpasses MPVR and ViLT by an average of $8.66\%$ and $9.85\%$, respectively, underscoring the effectiveness of our method in bolstering the model's resilience and performance across varying missing rate conditions.

Prompt learning enhances multimodal fusion, improving model performance. MPVR's prompting method lacks robustness, leading to overfitting and sensitivity to missing modality cases. Missing-aware ability alone is insufficient, necessitating more robust methods.
Our prompt exhibits modality-specificity and achieves missing-awareness through diverse fusion techniques. Multi-step prompt tuning aligns distinct modalities via adjustments, highlighting a trade-off between model performance and robustness.

\subsection{Ablation Study}
\label{ablation}


\begin{table}[!t]
    \centering
    \resizebox{\columnwidth}!{
        \begin{tabular}{c|c|c}
        \hline
        \multirow{2}{*}{\shortstack{Methods}} & 
        \multirow{2}{*}{\shortstack{Missing Rate \\ $\epsilon$}} &  
        \multirow{2}{*}{\shortstack{Hateful Memes  \\(AUROC)}}  \\
        && 
        \\
        \hline
        MuAP-w-tuning&\multirow{5}{*}{70\%}&\textbf{65.09}\\
        \cline{0-0} \cline{3-3}
        MuAP-w/o-single-stage&& 63.47\\
        \cline{0-0} \cline{3-3}
        MuAP-w/o-text-step&& 63.65\\
        \cline{0-0} \cline{3-3}
        MuAP-w/o-image-step&& 64.64\\
        \cline{0-0} \cline{3-3}
        MuAP-w/o-KL&&64.57\\
        \hline
        \end{tabular}
    }
     \caption{Ablation study to explore how multi-step prompt tuning improves model's performance. All models using the head-fusion strategy are trained and evaluated on missing-image scenarios.
    } 
    \label{table:ablation_multi_view}
\end{table}

\begin{figure}[!t]
    \centering
    \includegraphics[width=\columnwidth]{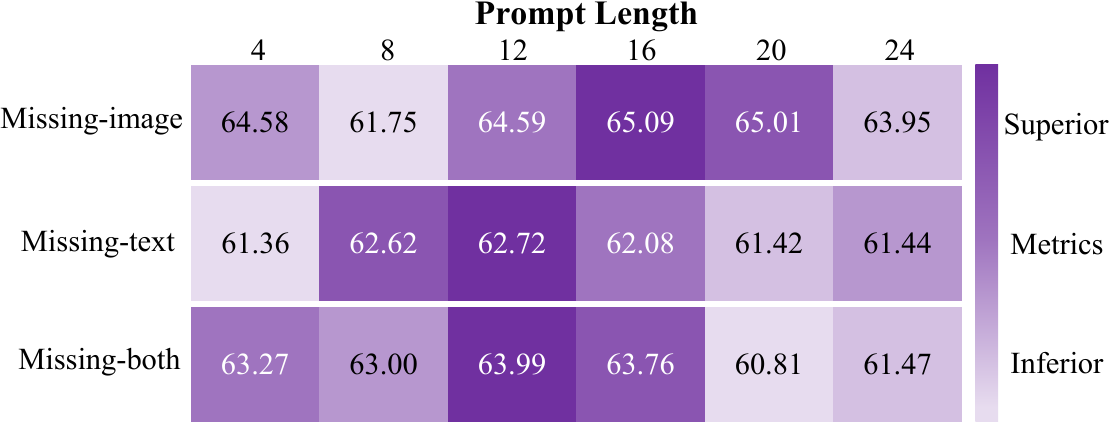}
    \caption{Ablation study on prompt length for head-fusion strategy. All models are trained and evaluated on various scenarios (e.g., missing-image) with $\epsilon$=70\%.}
    \label{fig:ablation_prompt_length}
\end{figure}

\paragraph{\textbf{Effectiveness of Multi-step Prompt Tuning}}
One of the most innovative aspects of our approach is the multi-step prompt tuning, consisting of single-stage and alignment-stage steps. We conducted experiments to assess the impact of it. As shown in Table~\ref{table:ablation_multi_view}, the variation with multi-step prompt tuning achieves the best performance, while the model without any tuning performs the worst. The experiment demonstrates that without iterative tuning steps, the model fails to capture crucial modality-specific information, which is essential for effective multimodal fusion. Other variations (e.g., removing text-step, KL divergence) also show different degrees of performance decrease, indicating that this module we set up to align modalities has a significant positive effect.

\paragraph{\textbf{Effectiveness of Prompt Length}}

In our proposed approach, the prompt length $L_P$ is a critical factor. For example, in the head-fusion prompting strategy, the final output length scales linearly with $(N_PL_P+L_V)$. Therefore, a judicious choice of $L_P$ is necessary to ensure computational efficiency and prevent information disruption during the training process.
We analyze the effect of prompt length in Figure~\ref{fig:ablation_prompt_length}. 
Consistent with intuition, model performance improves as prompt length $L_P$ increases, peaking at values between 12 and 16. This improvement can be attributed to the additional modal information provided at shorter lengths, preventing overfitting. However, a decline in performance is observed when the length exceeds 16. This observation indicates that excessively long prompts lead to a concatenation situation where the combined length nears the original embedding length, hindering effective learning.

\section{Conclusion}
In this paper, we have undertaken the pioneering effort to comprehensively investigate the robustness of prompt learning models when modalities are incomplete. Our experimental findings have revealed the high sensitivity of existing prompt learning models to the absence of modalities, resulting in substantial performance degradation. Building upon these insights, we propose a  Multi-step Adaptive Prompt Learning (MuAP) framework for missing-modality in the Vision-Language Model. 
We generate learnable modality-specific prompts and explore two prompt strategies to facilitate prompt learning in missing-modality Transformer models. To enable adaptive learning of multimodal prompts, we employ a multi-step tuning mechanism encompassing single-stage and alignment-stage tunings to perform multi-step modality alignments. This enables MuAP to acquire comprehensive knowledge from both modalities in a balanced manner. Extensive experiments conducted on benchmark datasets validate the effectiveness of MuAP. 

\section{Limitation}
First, due to time and computational constraints, we haven't tested our techniques on LLMs and larger datasets. Second, in our choice of modalities, we've focused solely on text and visuals using ViLT. It's crucial to incorporate additional modalities such as sound. It's essential for our proposed approach to demonstrate generalizability across diverse modalities, a focus for our upcoming work.
 Third, we have not explored more alignment methods due to the computational limitations. Finally, despite using few parameters, the overall  improvement is not substantial, but the robustness verification has significantly enhanced. Moving forward, more interpretable analysis will be carried out to comprehend the principles of the parameters' effects.
\appendix

\section{Implementation Details}
Regarding text modality, we use the bert-base-uncased tokenizer to tokenize our input sequence.
Depending on the dataset, the maximum length of text sentences is set differently. It is set to 128 for Hateful Meme, 512 for Food-101, and 1024 for MM-IMDB. For the image modality, following ~\cite{vit}, we extract 32 $\times$ 32 patches from the input image. Therefore, the input images are resized to 384 $\times$ 384 during the preprocessing stage.

For the missing situation, we follow~\cite{map} to keep the overall missing rate at $70\%$.
Considering various missing scenarios, we mainly set three  cases, including only the text modality (missing-text) or image modality (missing-imgae) missing $\epsilon\%$ while the other modality remains intact, and another type is both modalities (missing-both) are missing $\frac{\epsilon}{2}\%$ separately. The specific missing scenarios in training and inference experiments are shown in Table~\ref{table:baseline comparison}.

Moreover, the backbone parameters are initialized by pre-trained weights of ViLT. The length $L_p$ of learnable prompts is set to 16 by default in both head fusion and cross fusion. We set the maximum prompt layer number to 6 (i.e. the indices of layers to pre-pend prompts start from 0 and end at 5). 
The base learning rate is set at $1\times 10^{-2}$ using the AdamW optimizer~\cite{adamw} and weight decay at $2\times 10^{-2}$ to remain unchanged from ~ \cite{map}.
\section{Details of Various Datasets}

As previously mentioned, we have three distinct datasets: MM-IMDb~\cite{mmimdb}, UPMC Food-$101$~\cite{food101}, and Hateful Memes~\cite{hatememes}, each with its own objectives and evaluation metrics. 

\begin{figure}[h]
    \centering
    \includegraphics[width=\columnwidth]{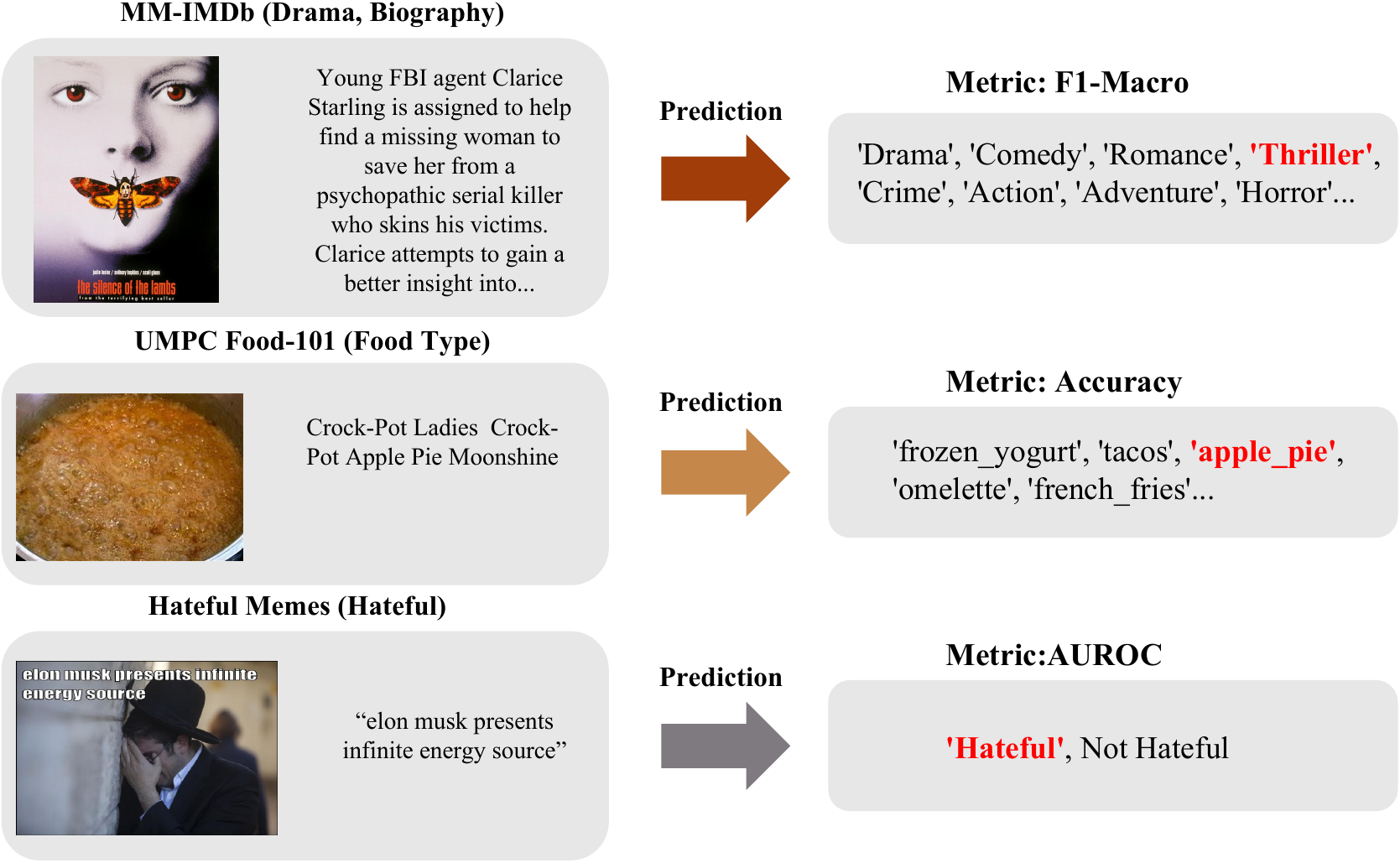}
    \caption{Detailed examples for three benchmark datasets.}
    \label{fig:datasets}
\end{figure}

To provide a clear overview of these datasets, Figure~\ref{fig:datasets} illustrates a comparison of their task objectives. MM-IMDb focuses on classifying movie genres, UMPC Food-$101$ is designed for food type classification, and Hateful Memes presents a formidable challenge in detecting hate speech across multiple modalities. As depicted in Figure \ref{fig:datasets}, the Hateful Memes dataset poses the greatest challenge due to its extensive composition of over $10,000$ newly generated multimodal instances. The intentional selection of these instances aims to pose difficulties for single-modal classifiers in accurately labeling them. For instance, a classifier relying solely on the text ``Elon Musk presents infinite energy source'' may not classify it as hateful. However, when accompanied by the corresponding image of Elon Musk placing his hand on his forehead, crucial contextual information is provided to identify its hateful connotation.
The tasks in MM-IMDb and UMPC Food-101 are notably less challenging due to explicit answers within the text. This is evident in the UMPC Food-$101$ example, where the classification result ``apple pie'' is directly mentioned in the text. Therefore, in our experimental setup, we primarily utilize the Hateful Memes dataset to effectively showcase the superiority of our approach compared to various baseline models.

\section{More Ablation Results}

\begin{table}[h]
    \centering
    \resizebox{\columnwidth}{!}{
        \begin{tabular}{c|cccc}
        \toprule
        \multirow{2}{*}{\shortstack{Missing-text  \\ Missing  rate $\epsilon$}} & 
        \multirow{2}{*}{ViLT } &
        \multirow{2}{*}{\shortstack{MPVR  \\(Input-level)}} &
        \multirow{2}{*}{\shortstack{MuAP \\ (Head Fusion)}} &    
        \multirow{2}{*}{\shortstack{MuAP \\ (Cross Fusion)}} \\ 
        \\
        \hline
        70\% &55.56 & \cellcolor{gray!60}\textbf{60.02} & 57.95 &\cellcolor{gray!20} 58.39\\
        \hline     
        50\%&59.63 &61.98 & \cellcolor{gray!60}\textbf{68.18} &\cellcolor{gray!20} 62.39\\
        \hline     
        30\%&65.47 & \cellcolor{gray!20} 67.80 & \cellcolor{gray!60}\textbf{68.17} & 64.85\\
        \hline
        10\%&66.37 &67.36 & \cellcolor{gray!60}\textbf{69.79} & \cellcolor{gray!20}69.65\\
        \bottomrule
        \end{tabular}
    }
     \caption{Ablation study of generalization ability on Hateful Memes. All models are evaluated on missing-text cases with different missing rates $\epsilon$. 
    } 
    \label{table:generalization_ability}
\end{table}

\paragraph{\textbf{Generalization Ability}}
Initially, we assume that real-world scenarios may involve missing modality instances due to device malfunctions or privacy concerns. However, the majority of existing datasets comprise modality-complete and meticulously annotated data. To address this inconsistency, we conducted experiments to investigate the impacts of a prompt learning model trained on complete modality datasets. 
In detail, all models are trained on complete modality cases and tested on scenarios with missing text at different rates.
In Table~\ref{table:generalization_ability}, our findings reveal that head-fusion and cross-fusion prompting exhibit robustness to this practical situation across numerous configurations. They consistently rank among the top performers, except for $\epsilon$ values of $70$\%. Our head-fusion prompting strategy exhibits remarkable performance, substantially enhancing both performance and robustness in the majority of scenarios, with an average AUROC of $66.02$\%, which is a $1.73$\% improvement compared to the average performance of MPVR ($64.29$\%). Meanwhile, the cross-fusion prompting strategy ranks second in most cases, showing a more pronounced sensitivity to specific settings compared to the head-fusion prompting strategy.
According to the findings elucidated in the paper, the cross-fusion prompting strategy proves to be effective in handling incomplete multimodal data, while the head-fusion prompting strategy exhibits exceptional robustness when dealing with complete multimodal data.

\begin{table}[h]
    \centering
    \resizebox{0.95\columnwidth}!{
        \begin{tabular}{c|c|c}
        \hline
        \multirow{2}{*}{\shortstack{Methods}} & 
        \multirow{2}{*}{\shortstack{Missing Rate \\ $\epsilon$}} &  
        \multirow{2}{*}{\shortstack{Hateful Memes  \\(AUROC)}}  \\
        && 
        \\
        \hline
        MuAP-w-tuning&\multirow{5}{*}{70\%}&\textbf{66.86}\\
        \cline{0-0} \cline{3-3}
        MuAP-w/o-single-stage&& 66.46\\
        \cline{0-0} \cline{3-3}
        MuAP-w/o-text-step&& 64.88\\
        \cline{0-0} \cline{3-3}
        MuAP-w/o-image-step&& 64.56\\
        \cline{0-0} \cline{3-3}
        MuAP-w/o-KL&&65.28\\
        \hline
        \end{tabular}
    }
     \caption{Ablation study to explore how multi-view prompt tuning improves model's performance. All models using the cross-fusion strategy are trained and evaluated on missing-image scenarios with missing rate $\epsilon$=$70$\%.  Best results in \textbf{bold}.
    } 
    \vspace{-2em}\label{table:ablation_multi_view_cross}
\end{table}

\paragraph{\textbf{Effectiveness Analysis in Cross-fusion}}

Due to space limitations in the main text, our analysis focused on assessing the effectiveness of the multi-view prompt tuning module with a head-fusion strategy. To attain a more profound comprehension of our pioneering multimodal alignment method, which encompasses multiple steps for enhancing understanding, we now evaluate its effectiveness using the cross-fusion prompting strategy. As depicted in Table \ref{table:ablation_multi_view_cross}, analogous to the preceding experimental findings, the model refined with multi-view prompting exhibits exceptional performance, surpassing all comparative models, while the untuned model performs the poorest. This validation evidence underscores the significance of iterative tuning in capturing modality-specific information that is pivotal for accomplishing successful multimodal fusion.

\begin{figure*}[!t]
    \centering
    \Huge
    \resizebox{0.9\linewidth}!{
    \begin{tabular}{cc}
    \begin{tabular}{c}\includegraphics[width=0.95\textwidth]{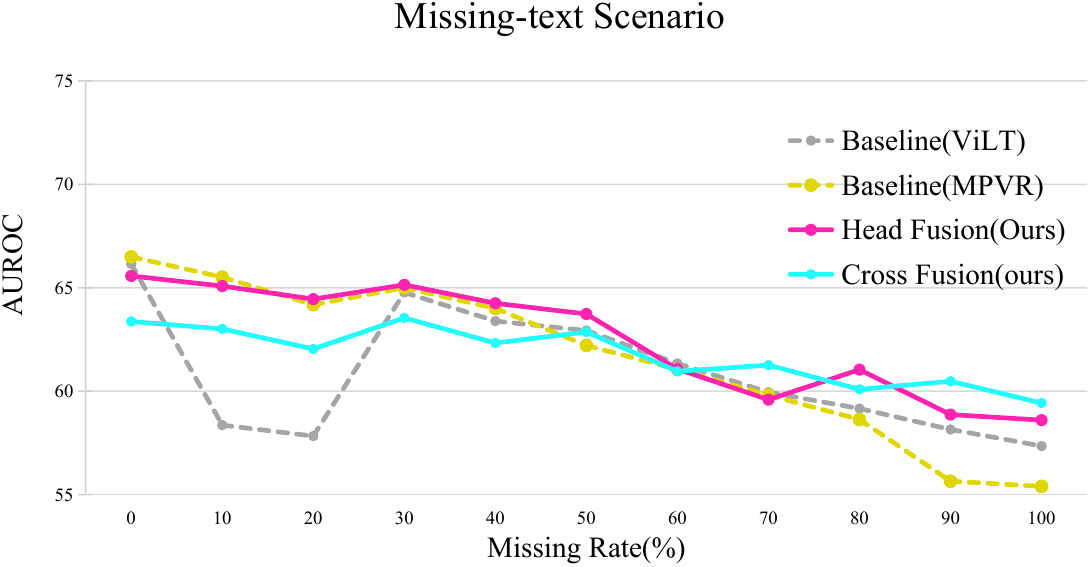}\end{tabular}
    &\begin{tabular}{c}\includegraphics[width=0.95\textwidth]{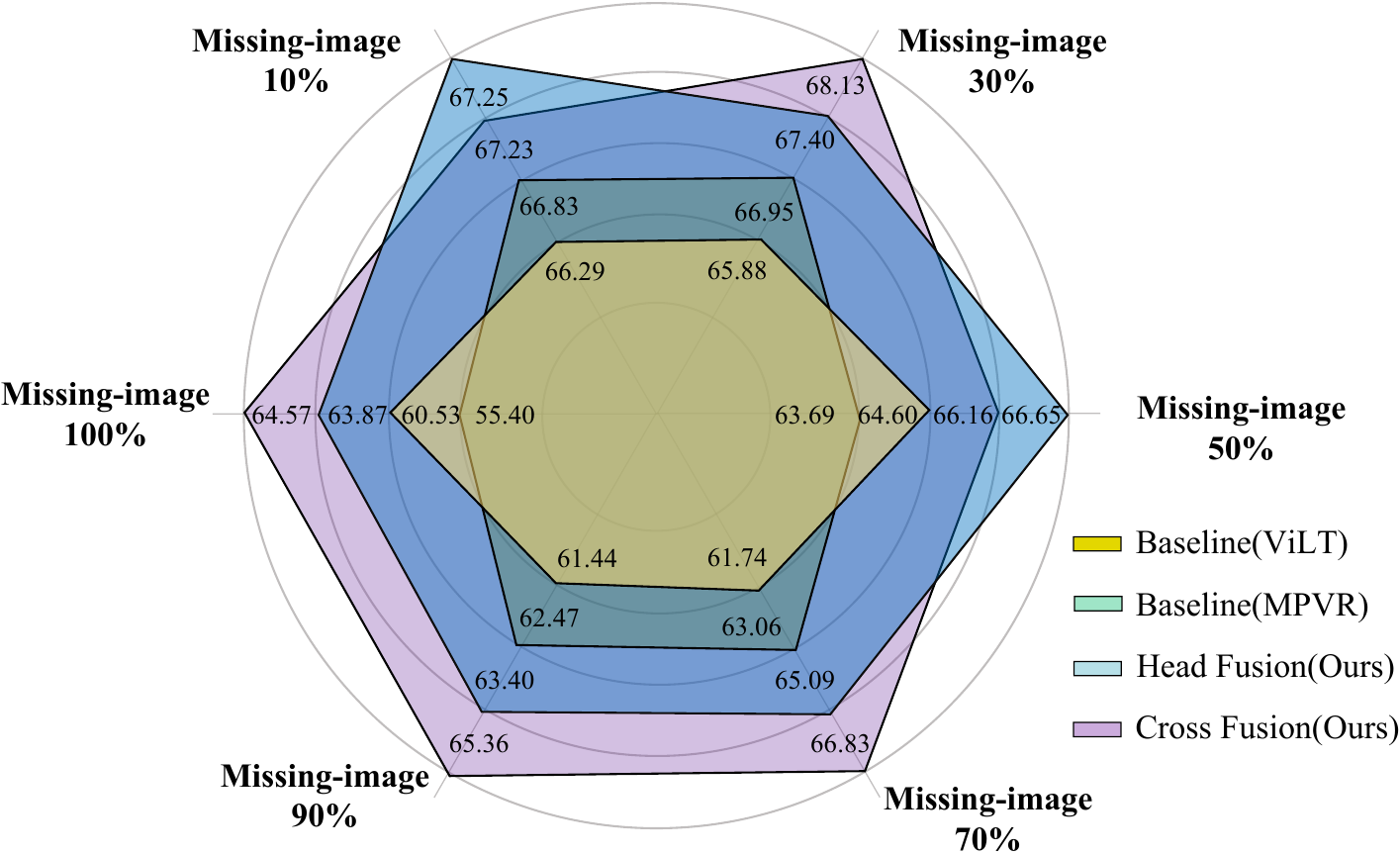}\end{tabular} \\
    (a) \textcolor{blue}{Train}: Missing-image 70$\%$; \textcolor{red}{Test}: Missing-text  
    &(b) \textcolor{blue}{Train}: Missing-image 70\%; \textcolor{red}{Test}: Missing-image 

    \end{tabular}
    }
    \caption{Robustness studies conducted by varying the missing rates in different evaluation scenarios for the Hateful Memes dataset. (a) Head-fusion models are trained using the missing-image scenario with $\epsilon$ = 70\%, and evaluations were performed on the opposite missing-text case.
    (b) All models are trained on the missing-image scenario with a 70\% missing rate, and tested on consistent cases with different missing rates, representing a transition from more complete data to less complete data.
    }
    \label{fig:ablation_different_missing_ratio_diff_train_test}
    \vspace{-1mm}
\end{figure*}

\paragraph{\textbf{Robustness to Different Missing Settings}}
We conduct experiments with different missing scenarios to demonstrate our method's robustness across various scenarios during the training and testing process. We aim to showcase the effectiveness of our method in improving both performance and robustness.

In previous work, ~\cite{maple,liu2024modality} use textual modality as the main modality. So we evaluated models trained on a missing-image scenario with a $70\%$ missing rate and diverse missing-text scenarios with varying $\epsilon$ values. Figure~\ref{fig:ablation_different_missing_ratio_diff_train_test}(a) shows that head-fusion consistently outperforms MPVR across scenarios, with our proposed strategies remaining robust even with increasing missing rates, achieving average AUROC values of $62.49\%$ and $61.76\%$ for head-fusion and cross-fusion, respectively.
Our approach maintains stable performance even in highly challenging scenarios with higher missing rates, unlike MPVR, which becomes ineffective when the missing rate surpasses $80\%$. We attribute this improvement to our $f_{\mathsf{missing}}$ function, implemented using residual connections, familiarizing the model with complete and missing data scenarios, effectively facilitating information supplementation.

Figure~\ref{fig:ablation_different_missing_ratio_diff_train_test}(b) illustrates the robustness of our proposed method, as it consistently outperforms MPVR, particularly when tested with varying missing rates while maintaining consistent missing-image settings during training. Our multimodal prompts effectively tackle missing-awareness and modality-specificity,  significantly boosting the robustness of prompt learning.

\begin{figure}[t]
    \includegraphics[width=1\columnwidth]{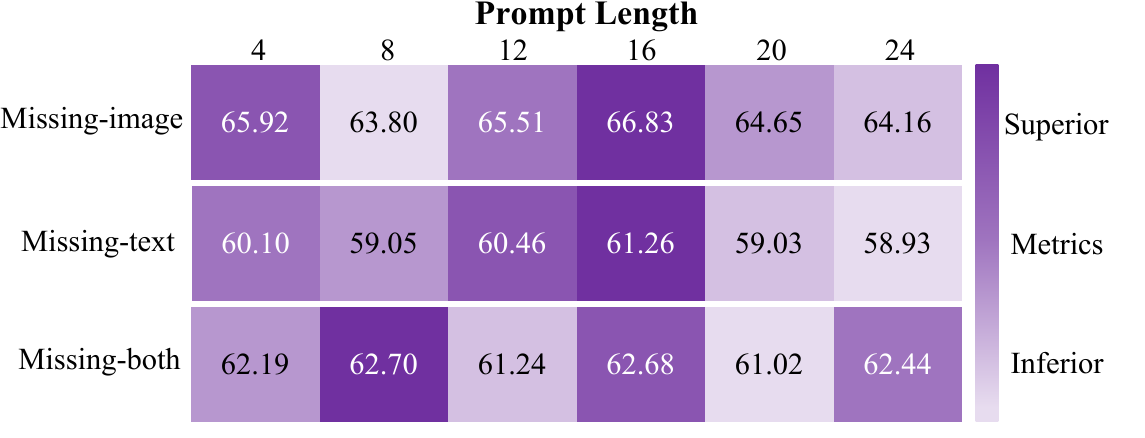}
    \caption{
    Ablation study on prompt length for cross-fusion
strategy. All models are trained and evaluated on various
scenarios (e.g., missing-image, missing-text) with $\epsilon$=70\%.
    }
    \label{fig:ablation_cross_pl}
    \vspace{-1mm}
\end{figure}

Moreover, we have analyzed the model performance in various prompt lengths with the head-fusion strategy in the main text. However, in the proposed cross-fusion prompting approach, the output length exhibits a linear increase, directly proportional to the sum of $(2N_PL_P + L_V)$. This linear growth becomes notably more substantial as the length of the prompt escalates compared to head-fusion, which may lead to increased computational demands and potential efficiency challenges. 
The analysis presented in Figure~\ref{fig:ablation_cross_pl} reveals a distinct trend compared to head-fusion. It is noteworthy that the top-3 performances in each scenario exhibit variability. Notably, when $L_P$ ranges from 4 to 8, the performance is consistently strong, achieving the highest AUROC of 62.70\% in the missing-both case. This suggests that even smaller values of $L_P$ can yield excellent performance. However, akin to head-fusion, the optimal performance is observed when $L_P$ approaches 16. These findings suggest that our cross-fusion method is particularly sensitive to the prompt length due to the rapid accumulation of sequence length, potentially leading to overfitting and inefficient computation.

\begin{table}[h]
\caption*{\textbf{Hateful Memes}}
\centering
\resizebox{0.35\textwidth}{!}{%

\begin{tabular}{c>{\columncolor{gray!20}}cccccc}
	    \toprule
\diagbox{$\lambda_v$}{$\lambda_t$}		&0.4		&0.5		&0.6		&0.7		&\multicolumn{1}{|c}{AVG}\\
    \midrule
        \rowcolor{gray!20}
        0.3	&65.09	&64.44			&65.50	&66.01	        &\multicolumn{1}{|c}{65.26}
        \\
        0.4	&64.88	&63.90			&64.31	&64.06	        &\multicolumn{1}{|c}{64.29}\\
        0.5	&64.29	&65.04			&64.40	&63.35        &\multicolumn{1}{|c}{64.27}\\
        0.6	&64.08	&64.11		&64.03	&63.55	        &\multicolumn{1}{|c}{63.94}\\
    \midrule
        AVG	&64.59	&64.37		&64.56	&64.24&\multicolumn{1}{|c}{64.44}	\\
    \bottomrule
    \end{tabular}
}

\captionof{table}{Hyperparameters selection analysis on the hyper-parameter $\lambda$ for both modalities with Hateful Memes~\cite{hatememes}. All head-fusion models are trained and tested on missing-image scenario with $\epsilon$=$70$\%}\label{lambdaeffect}
\end{table}

\section{The Selection of Multi-view Prompt Tuning 
Hyperparameters}

To demonstrate our selection of the involved hyperparameters, we further analyze the impact of the hyperparameters $\lambda_t$ and $\lambda_v$ with Hatefull Memes~\cite{hatememes}. From Table~\ref{lambdaeffect}, it can be seen that when $\lambda_t$ is $0.4$, the overall performance is the best, and the same situation occurs when $\lambda_v$ is $0.3$. Therefore, in the remaining experiments,  we maintain $\lambda_t$ at $0.4$ and $\lambda_v$ at $0.3$ to gain the maximum performance.

\end{document}